\documentclass[letterpaper, 10 pt, conference]{ieeeconf}  

\IEEEoverridecommandlockouts                              

\usepackage{amsmath}
\usepackage[dvipsnames]{xcolor}
\usepackage{algorithm}
\usepackage{algpseudocode}
\usepackage{graphicx}
\usepackage{amssymb}
\usepackage{cite}
\usepackage{booktabs}
\usepackage{multirow}
\usepackage[pdfa,colorlinks,bookmarksopen,bookmarksnumbered,allcolors=blue]{hyperref}

\title{\LARGE \bf
MAPLE: Mondrian-Anchored Prior for Learning-free Indoor Exploration
}

\author{
Kelvin Aladum, Nisar Ahmed
}

\begin{document}

\maketitle
\thispagestyle{empty}
\pagestyle{empty}


\begin{abstract}
Robots exploring unknown indoor environments must decide where to move while much of the environment remains unseen. Predicting the occupancy of unseen space beyond the current sensor range can improve these decisions, but recent approaches commonly rely on learned models that require offline training and repeated inference during exploration. We introduce the Mondrian-Anchored Prior for Learning-free Indoor Exploration (MAPLE), a training-free predictive approach that adapts the Mondrian process into an online generative prior conditioned on observed geometric cues in the current partial map. MAPLE generates multiple plausible map hypotheses and evaluates visibility separately within each to estimate frontier gain, preserving differences in predicted visibility across hypotheses. Across KTH floor-plans and real-SLAM environments at multiple sensor ranges, MAPLE consistently improves over non-predictive exploration and remains competitive with learned predictive approaches. Its lightweight procedural construction enables predictive exploration with limited computational resources. Experimental results further show that more accurate map prediction does not necessarily produce better exploration, highlighting the importance of how predictions are represented and used in exploration decisions.
\end{abstract}


\section{Introduction}
\label{sec:intro}
Autonomous exploration is a key capability for robots operating in unknown or dangerous environments, with applications in search and rescue, emergency response, and infrastructure inspection \cite{wang2018collaborative,luperto2020robot}. In these settings, robots must repeatedly decide where to move while much of the environment remains unseen. Frontier-based exploration makes this decision by identifying frontiers at the boundary between observed free space and unknown space \cite{yamauchi1997frontier}. Connected frontier regions are represented by frontier candidates that serve as possible exploration goals, and selecting among them directly affects exploration efficiency.

Classical frontier methods select frontiers using measures such as travel cost, expected coverage, or information gain \cite{yamauchi1997frontier,gonzalez2002navigation,julian2014mutual,charrow2015information}. These measures are typically computed from the current partial occupancy map, so they provide little information about occupancy beyond the sensor range, even though that occupancy can strongly affect how much new space a frontier will reveal. Predictive exploration addresses this limitation by estimating the occupancy of unseen space before frontier selection \cite{shrestha2019learned,georgakis2022uncertainty,ericson2024beyond,ho2025mapex,tan20264cnet,harutyunyan2025mapexrl,baek2025pipe}.
Most recent predictive approaches rely on learned models \cite{georgakis2022uncertainty,ho2025mapex,tan20264cnet}. These methods can produce useful occupancy predictions but require offline training and repeated inference during exploration. Their computational demands can be restrictive for robots with limited compute and power, and performance may degrade when the operating environment differs from the training data. This tradeoff is particularly relevant at short sensor ranges, where frontier selection depends more heavily on occupancy beyond the current sensor range \cite{ericson2021understanding}. These considerations motivate a lightweight alternative that can generate predictions directly from the partial map during exploration.

\begin{figure}[t]
  \centering
  \includegraphics[width=1\columnwidth]{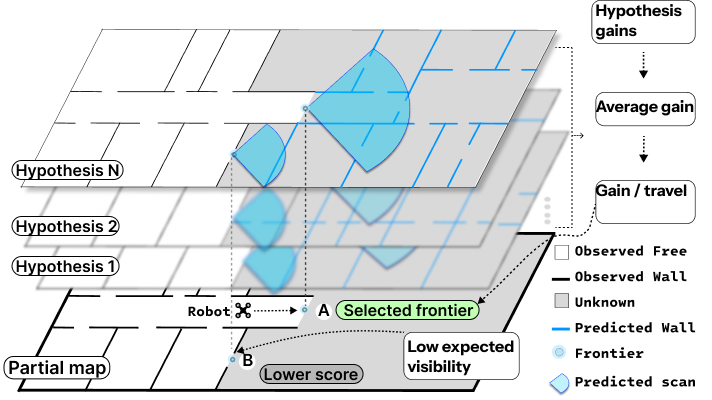}
  \caption{MAPLE evaluates frontier candidates over multiple plausible map completions, treated as map hypotheses. Different predicted walls can give the same candidate different visibility across hypotheses. The resulting gains are averaged before accounting for travel cost and selecting the next frontier candidate.}
  \label{fig:overview}
\end{figure}

We introduce the \textbf{M}ondrian-\textbf{A}nchored \textbf{P}rior for \textbf{L}earning-free Indoor \textbf{E}xploration (MAPLE), a training-free predictive approach for two-dimensional occupancy-grid exploration. MAPLE builds on the Mondrian process \cite{roy2008mondrian}, which recursively partitions a rectangular region using axis-aligned cuts, and guides this process using geometric cues from the current partial map. Repeated sampling produces multiple plausible map completions of the unseen space, each treated as a map hypothesis.
Different map hypotheses can imply different visibility from the same frontier candidate. MAPLE therefore evaluates each hypothesis separately and averages the resulting coverage gains before accounting for travel cost. Fig.~\ref{fig:overview} illustrates how differing predicted walls can change visibility for the same candidate.

We evaluate MAPLE on ten KTH floor plans and ten real-SLAM maps at sensor ranges of 5, 10, and 20 m. The experiments examine end-to-end exploration, map-prediction quality, frontier-decision quality, and computational cost. Controlled ablations also test the effects of the map-prediction method, gain estimator, conditioning on observed geometry, and the number of hypotheses.

Our contributions are as follows:
\begin{itemize}
\item We introduce MAPLE, a lightweight training-free generative prior that adapts the Mondrian process to generate multiple plausible map hypotheses from geometric cues in the current partial map.
\item We formulate a hypothesis-wise frontier-gain estimator that evaluates visibility separately within each map hypothesis before averaging the resulting gains.
\item We provide a controlled comparison of learned, training-free, and non-predictive exploration under shared sensing, frontier candidates, and travel cost, separating prediction, decision, exploration, and computational performance.
\end{itemize}
The results show that predictive exploration depends not only on the quality of map predictions but also on how those predictions are used during frontier selection. More accurate map prediction alone does not necessarily lead to more efficient exploration.


\section{Related Work}

\textit{Frontier and Information-Theoretic Exploration:} Classical frontier-based exploration methods provide established ways to prioritize frontiers but typically only reason from the current partial map. Early work prioritizes nearby frontiers, while later work balances expected visible area against travel cost \cite{yamauchi1997frontier,gonzalez2002navigation}. Information-theoretic approaches instead score actions by expected information gain or mutual information \cite{julian2014mutual,charrow2015information}. Receding-horizon planners and hierarchical planners extend these methods to much larger and three-dimensional environments \cite{bircher2016receding,cao2021tare}. MAPLE retains this frontier-selection framework but uses multiple map completions to represent unseen occupancy beyond the current partial map.

\textit{Learned Prediction of Unseen Space:} Predictive exploration addresses this limitation by estimating occupancy beyond the current sensor range. Earlier work used previously mapped environments to reason about unseen space \cite{strom2015predictive}. Learned approaches later completed or anticipated unseen occupancy for exploration and navigation \cite{shrestha2019learned,katyal2019uncertainty,ramakrishnan2020occupancy,wei2021inpainting}. UPEN uses an ensemble of occupancy predictors to guide exploration toward uncertain regions \cite{georgakis2022uncertainty}. Ericson and Jensfelt predict unseen walls as line segments \cite{ericson2024beyond}. MapEx combines LaMa map completion with prediction uncertainty, and SEER learns to predict information gain for aerial exploration \cite{ho2025mapex,suvorov2022resolution,tao2023seer}. More recent methods use conditional consistency or diffusion models for unseen occupancy, including 4CNet and SceneSense \cite{tan20264cnet,reed2024scenesense}. Achey et al. further integrate generative occupancy synthesis into real-world exploration \cite{achey2026generative}. These methods demonstrate the value of predicting unseen occupancy but rely on learned models and repeated inference. In contrast, MAPLE generates map completions online without offline training.

\textit{Structural and Training-Free Priors:} Indoor spaces contain geometric regularities that can support reasoning about unseen space. Aydemir et al. showed from over 38,000 rooms that room geometry and connectivity exhibit useful regularities \cite{aydemir2012can}. Luperto et al. used prior floor plans to facilitate exploration and later predicted the remaining geometry of partially observed rooms from the current occupancy map \cite{luperto2020robot,luperto2021exploration}. Related work on building and floor plan generation has used procedural modeling, stochastic layout synthesis, and learned generative models \cite{muller2006procedural,merrell2010residential,shabani2023housediffusion}. These methods primarily target layout generation or rely on prior maps. MAPLE instead uses a Mondrian-based generative process \cite{roy2008mondrian} conditioned on the current partial map to sample plausible map completions online.

\textit{Multi-hypothesis Beliefs and Exploration Decisions:} When multiple map hypotheses are available, they must be combined or evaluated to guide exploration. Prior work has addressed this mainly in the context of SLAM uncertainty. Stachniss et al. evaluate exploration actions over map and pose hypotheses represented by a Rao-Blackwellized particle filter \cite{stachniss2005information}. Wang and Englot combine utilities across hypotheses induced by ambiguous data association \cite{wang2018robust}. Recent learned approaches also summarize their ensembles before planning. UPEN converts ensemble disagreement into a per-cell uncertainty map \cite{georgakis2022uncertainty}. MapEx computes frontier scores from the ensemble mean and pixel-wise variance \cite{ho2025mapex}. CogniPlan plans with multiple layouts produced by a conditional generative model and couples them to a learned graph-based planner \cite{cogniplan2025}. Ericson et al. show that planner behavior can strongly affect the benefit obtained from map prediction \cite{ericson2021understanding}. More broadly, planning-aware evaluation has shown that prediction accuracy can be poorly aligned with downstream decision quality \cite{ivanovic2022injecting}. Recent exploration work finds that improved information-gain estimates do not necessarily improve exploration efficiency \cite{ericson2025information}. MAPLE isolates this issue within a common frontier-selection framework by preserving individual map hypotheses during visibility evaluation and testing how this choice affects exploration.


\section{Problem Formulation}
\label{sec:setup}
At each replanning step, the problem is selecting the reachable frontier candidate with the highest predicted gain-to-travel score despite uncertainty about unseen space.
\subsection{Exploration Task}
\label{sec:setup:task}
The environment is a bounded two-dimensional space represented by a ground-truth occupancy grid \(M\), whose cells are free or occupied and which is unknown to the robot. At time \(t\), the exploration planner receives the robot pose estimate \(x_t\) and partial occupancy map \(M_t\) from an upstream localization and mapping system. Cells in \(M_t\) are labeled observed free, observed occupied, or unknown, and \(U_t\) denotes the set of unknown cells.

Frontier cells are observed free cells adjacent to \(U_t\), and connected frontier cells form frontier regions \cite{yamauchi1997frontier}. Each reachable frontier region is represented by the frontier cell closest to its centroid, and these representative cells form the set of frontier candidates \(\mathcal F_t\). A map completion \(M'\) is a binary occupancy map that preserves the observed cells of \(M_t\) and assigns every cell in \(U_t\) as either free or occupied.

The sensor is characterized by maximum range \(R\) and field of view \(\Phi\). For a limited-field-of-view sensor, each frontier candidate also specifies a sensing heading. For \(f \in \mathcal F_t\) and map completion \(M'\), let \(V(f,M')\) denote the cells visible from \(f\) by raycasting through \(M'\) within field of view \(\Phi\) and maximum range \(R\). Each ray terminates at the first occupied cell or the range limit. The coverage gain is
\begin{equation}
  g_t(f, M') = \bigl|\, V(f, M') \cap U_t \,\bigr|,
  \label{eq:gain}
\end{equation}
where \(|\cdot|\) denotes set cardinality. Thus, \(g_t(f,M')\) is the number of cells in \(U_t\) visible from \(f\) under \(M'\).

\subsection{Shared Frontier Selection Rule}
\label{sec:setup:rule}
Given the current partial map \(M_t\) and frontier candidate set \(\mathcal F_t\), the planner selects the next frontier candidate by maximizing the estimated gain-to-travel score:
\begin{equation}
  f_t^{\star} \;=\; \arg\max_{f \in \mathcal{F}_t} \; \frac{\widehat{G}_t(f)}{c_t(x_t, f)}
  \label{eq:rule}
\end{equation}
Here, \(f_t^\star\) is the selected frontier candidate, \(\widehat{G}_t(f)\) is its estimated frontier gain, and \(c_t(x_t,f)>0\) is the length of the shortest path from the current robot pose estimate \(x_t\) to \(f\) through observed free space in \(M_t\). Since \(\mathcal F_t\) contains only reachable frontier candidates, this cost is finite. If \(\widehat{G}_t(f)\) is the same positive constant for all candidates, Eq. ~(\ref{eq:rule}) reduces to nearest-frontier selection under the same path cost.

\subsection{Gain Estimation from Multiple Map Predictions}
\label{sec:setup:estimators}
Map-prediction methods may produce multiple occupancy predictions of the unseen space. Let \(P_t^{(1)},\ldots,P_t^{(N)}\) denote these predictions, where \(P_t^{(i)}(u)\in[0,1]\) is the predicted occupancy of cell \(u\). For MAPLE, the \(N\) predictions are binary map hypotheses \(M_t^{(i)}\), so \(P_t^{(i)}(u)\in\{0,1\}\). We consider two ways of using multiple predictions to estimate the frontier gain \(\widehat{G}_t(f)\) in Eq.~(\ref{eq:rule}).

The aggregated estimator, following MapEx \cite{ho2025mapex}, computes the cell-wise mean prediction \(\bar{P}_t\) and variance \(\sigma_t^2(u)\) across the \(N\) predictions. Probabilistic raycasting on \(\bar{P}_t\) produces the estimated visible region \(\bar{V}_t(f)\), giving:
\begin{equation}
  \widehat{G}_t^{\mathrm{agg}}(f) \;=\; \sum_{u \in \bar{V}_t(f) \cap U_t} \sigma_t^2(u).
  \label{eq:igmean}
\end{equation}

The hypothesis-wise estimator applies to binary map hypotheses \(M_t^{(1)},\ldots,M_t^{(N)}\). It evaluates visibility separately in each hypothesis and averages the resulting coverage gains:
\begin{equation}
  \widehat{G}_t^{\mathrm{hyp}}(f) \;=\; \frac{1}{N} \sum_{i=1}^{N} g_t\bigl(f, M_t^{(i)}\bigr).
  \label{eq:igps}
\end{equation}

\begin{figure}[t]
  \centering
  \includegraphics[width=0.99\columnwidth]{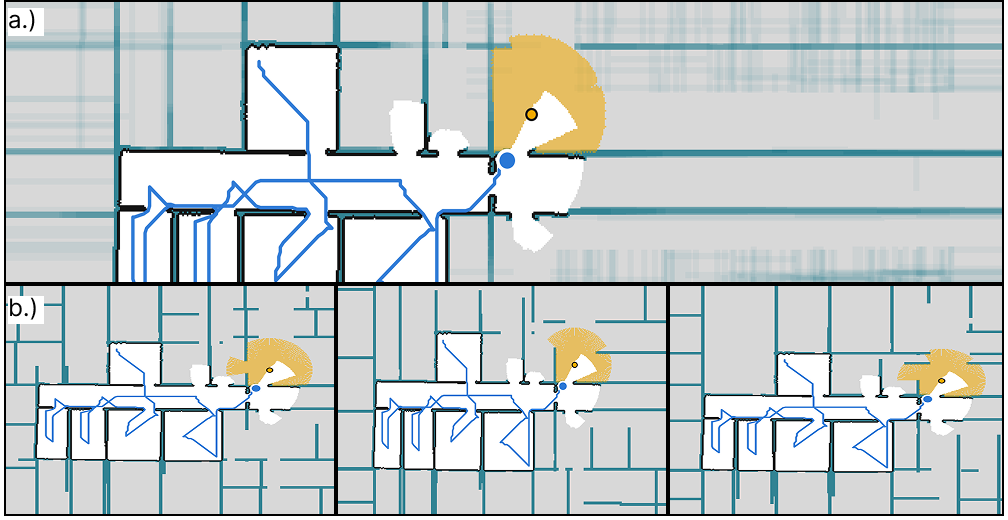}
    \caption{Effect of aggregation on predicted visibility for the same frontier candidate in a real-SLAM map. (a) Raycasting on the mean occupancy prediction, where aggregation changes predicted occupancy boundaries and the resulting visible region. (b) Raycasting the same frontier candidate independently on three map hypotheses, whose different occupancy predictions produce different visibility before the gains are averaged.}
  \label{fig:mean_vs_sample}
\end{figure}

The aggregated estimator combines predictions before visibility is evaluated, whereas the hypothesis-wise estimator preserves each binary map hypothesis during visibility evaluation and averages only the resulting gains. Fig.~\ref{fig:mean_vs_sample} illustrates this distinction.

\section{MAPLE}
\label{sec:method}
\begin{figure*}[t]
  \centering
  \includegraphics[width=1\textwidth]{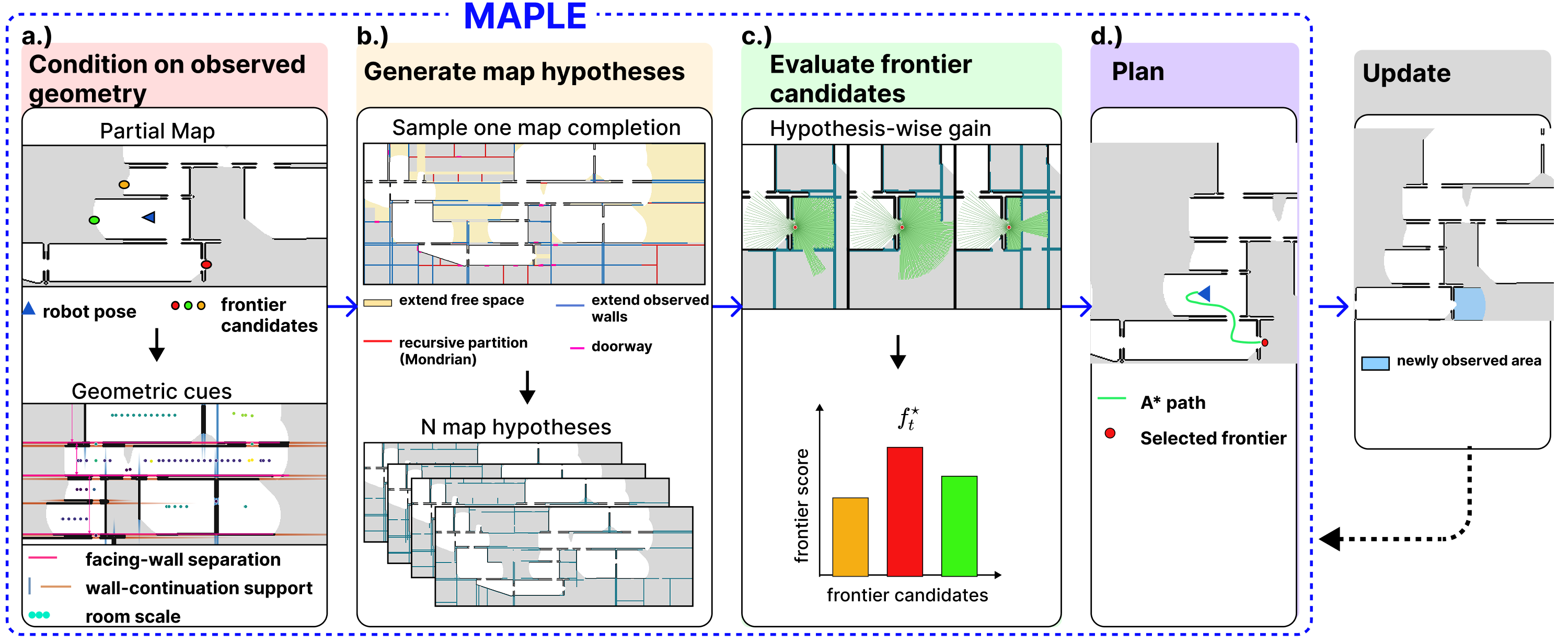}
  \caption{Overview of one MAPLE replanning step. (a) Geometric cues are extracted from the current partial map \(M_t\) to condition the generative prior. (b) MAPLE samples \(N\) map hypotheses by extending observed walls and free space, applying the observation-conditioned Mondrian partition, and restoring connectivity. (c) Each frontier candidate is evaluated independently in each hypothesis, the resulting gains are averaged, and the gain-to-travel score is used to select \(f_t^\star\). (d) An A* path is planned from the current robot pose to the selected frontier candidate. After execution, new observations update the partial map and initiate the next replanning step.}
  \label{fig:pipeline}
\end{figure*}

MAPLE uses a stochastic construction that induces a conditional distribution \(p_{\mathrm{MAPLE}}(M' \mid M_t)\) over map completions, which serves as its training-free generative prior. At each replanning step, MAPLE extracts geometric cues from the current partial map \(M_t\) and draws \(N\) conditionally independent map hypotheses \(M_t^{(i)} \sim p_{\mathrm{MAPLE}}(\cdot\mid M_t)\). Each hypothesis preserves the observed cells of \(M_t\) and predicts occupancy only in \(U_t\). MAPLE evaluates these hypotheses using the hypothesis-wise gain estimator in Eq.~(\ref{eq:igps}), as summarized in Fig.~\ref{fig:pipeline} and Algorithm~\ref{alg:maple}.

\subsection{Mondrian Process}
\label{sec:method_prior}

The Mondrian process recursively partitions a bounded rectangular region using axis-aligned cuts \cite{roy2008mondrian}. For a rectangular region \(B\) with width \(w\), height \(h\), and remaining budget \(\lambda\), the cost \(E\) of the next cut is sampled from an exponential distribution with rate \(w+h\):
\begin{equation}
E \sim \mathrm{Exp}(w + h).
\label{eq:mondrian_interval}
\end{equation}

If \(E>\lambda\), \(B\) remains intact. Otherwise, the \(x\)-axis is selected with probability \(w/(w+h)\) and the \(y\)-axis with probability \(h/(w+h)\), producing a vertical or horizontal cut, respectively. The cut coordinate is sampled uniformly along the selected axis, and the process recurses independently on the two resulting regions with remaining budget \(\lambda-E\). Regions with larger \(w+h\) therefore tend to receive more cuts, while the axis with greater extent is more likely to be selected.
MAPLE retains the recursive partitioning and axis-selection rule of the Mondrian process but conditions it on observed geometry in the current partial map. In particular, it replaces the budget-based stopping rule and uniform cut-location distribution with observation-conditioned rules. We refer to this conditioning as anchoring. The observed geometry influences how walls and free space are extended into \(U_t\), when partitioning stops, and where new cuts are placed.

\subsection{Conditioning on Observed Geometry}
\label{sec:method_conditioning}

Let \(\mathcal L\) denote the axis-aligned wall lines extracted from \(M_t\). The wall-continuation support fields \(L_v\) and \(L_h\) represent likely continuations of observed vertical and horizontal walls. Support is strongest along observed wall coordinates and decays beyond their endpoints, with longer walls retaining support farther into \(U_t\).
The local room-scale field \(\hat r(y,x)\) is estimated from nearby cross-sectional widths of observed free space. A map-level room scale \(\bar r_t\) summarizes observed parallel-wall separations and repeated-wall spacing, with observed free-space width providing a lower bound. It parameterizes wall-continuation length and provides a fallback when local room-scale evidence is unavailable.
For each wall line \(\ell\in\mathcal L\), the facing-wall measurements \(D_\ell\) contain pairs \((d,\kappa)\), where \(d\) is the separation to an observed parallel wall and \(\kappa\in[0,1]\) measures the consistency of that separation along \(\ell\). Larger \(\kappa\) provides stronger support for continuing the same separation into \(U_t\).

\subsection{Sampling Map Extensions}
\label{sec:method_sampling}
To generate each map hypothesis \(M_t^{(i)}\), MAPLE first extends observed walls and free space into \(U_t\) before partitioning the remaining unknown cells. For an observed wall line \(\ell\) entering \(U_t\), MAPLE samples its continuation length from a geometric distribution whose mean depends on the observed length of \(\ell\) and \(\bar r_t\). The continuation stops when it would cross observed free space or reach another wall. For a facing-wall measurement \((d,\kappa)\in D_\ell\), a parallel continuation at separation \(d\) is added with probability \(\alpha_f\kappa\), where \(\alpha_f\in[0,1]\) is fixed.
Extending walls alone can incorrectly close a corridor or doorway that continues into \(U_t\). MAPLE therefore also extends observed free space into \(U_t\). For each hypothesis, \(J\sim\mathrm{LogNormal}(0,\sigma_J^2)\) varies the extension length, with greater observed free-space width allowing farther extension. Extension stops at walls or where strong perpendicular wall-continuation support indicates a likely wall. The resulting predicted free space discourages subsequent cuts from passing through it.

As part of generating each map hypothesis, MAPLE samples a plausible building extent. Let \(B_t^{\mathrm{obs}}\) denote the axis-aligned bounding box of the observed free and wall cells in \(M_t\). Let \(\mu_{f,t}\) and \(\sigma_{f,t}\) be the mean and standard deviation of the log lengths of fully observed free-space spans bounded by walls at both ends. For each of the four map directions \(k\in\{\mathrm{up},\mathrm{down},\mathrm{left},\mathrm{right}\}\), MAPLE identifies observed free-space runs that reach unknown cells in that direction and sets \(w_k^\star\) to the longest such run. MAPLE then draws

\begin{equation}
    \begin{aligned}
        \tilde w_k&\sim\operatorname{LogNormal}(\mu_{f,t},\sigma_{f,t}^2) \ | \ \tilde w_k>w_k^\star, \\ \Delta_k &=\tilde w_k-w_k^\star .
    \end{aligned}
\end{equation}
The residual \(\Delta_k\) extends the selected free-space run into unseen space. The sampled building extent is obtained by expanding \(B_t^{\mathrm{obs}}\) to include this sampled reach.

\begin{algorithm}[t]
\caption{One replanning step of MAPLE}
\label{alg:maple}
\begin{algorithmic}[1]
\Require partial map $M_t$, pose $x_t$, number of hypotheses $N$
\State $ \mathcal F_t \gets$ frontier candidates of $M_t$
\For{$i = 1, \dots, N$}
    \State $M_t^{(i)} \sim p_{\mathrm{MAPLE}}(\cdot \mid M_t)$
\EndFor
\For{each $f \in \mathcal F_t$}
    \State $\widehat{G}_t^{\mathrm{hyp}}(f)
    \gets \frac{1}{N}\sum_{i=1}^{N}
    g_t\bigl(f,M_t^{(i)}\bigr)$
    \Comment{Eq.~\eqref{eq:igps}}
\EndFor
\State \Return $f_t^{\star}
    = \arg\max_{f \in \mathcal F_t}
    \widehat{G}_t^{\mathrm{hyp}}(f)/c_t(x_t,f)$
    \Comment{Eq.~\eqref{eq:rule}}
\end{algorithmic}
\end{algorithm}

\subsection{Observation-Conditioned Recursive Partition}
\label{sec:method_partition}

Given the sampled wall and free-space extensions, MAPLE recursively partitions the map, with new wall segments restricted to the remaining unknown cells. At each recursive step, \(B\) denotes the current rectangular region being partitioned. MAPLE places the next cut along an observed or continued wall if the wall spans at least a fraction \(\rho_s\) of \(B\)'s extent along the wall axis and lies at least \(d_s\) from its boundary.
If no wall satisfies this condition, MAPLE uses the local room scale at the center of \(B\),
\( r_B=\hat r(y_B,x_B), \) to determine whether further partitioning is needed.
Recursion stops when \(w<2r_B\) and \(h<2r_B\). Candidate cuts are restricted so that both resulting regions remain at least \(r_B\) wide along the selected axis. When both axes are feasible, MAPLE uses the Mondrian axis-selection rule from Sec.~\ref{sec:method_prior}.

Let \(\xi\) denote the cut coordinate along the selected axis. MAPLE samples
\begin{equation}
\begin{aligned}
    \xi &\sim q_B^{(i)},\\ q_B^{(i)}(\xi)\propto \big[\psi_d(\xi)+\psi_L(\xi) &+ \psi_r(\xi)+\psi_p(\xi)\big]\, \psi_{\mathrm{wr}}^{(i)}(\xi), 
    \end{aligned}
\end{equation}

Here, \(\psi_L\) favors coordinates with strong wall-continuation support and \(\psi_d\) favors cuts near observed geometry and predicted free space. When wall-continuation support is weak, \(\psi_r\) favors cuts at repeated multiples of the local room scale and \(\psi_p\) favors repeated-wall spacing. Finally, \(\psi_{wr}^{(i)}\) downweights cuts through observed or predicted free space.

\subsection{Validity and Connectivity}
\label{sec:method_connectivity}
Because cuts are not rendered through predicted free space, one cut can produce several wall segments. A segment is retained only when both endpoints are grounded by observed or continued walls, observed or predicted free space, the boundary of the sampled building extent, or another retained segment. MAPLE iteratively prunes unsupported segments until all remaining segments satisfy this condition.

The retained walls are rendered using the wall thickness estimated from \(M_t\). If predicted free-space regions remain disconnected from observed free space, MAPLE restores connectivity by opening a doorway through a predicted wall without modifying observed occupied cells. Doorway positions are sampled from doorway locations observed along walls in \(M_t\), or uniformly along the wall if none has been observed. The resulting binary occupancy map is the map hypothesis \(M_t^{(i)}\).


\section{Experiments}
\label{sec:experiments}

We evaluate MAPLE on end-to-end exploration, map-prediction quality, frontier-decision quality, and computational cost. Controlled ablations then isolate the effects of the map-prediction method, gain estimator, geometric conditioning, and the number of hypotheses.

\subsection{Experimental Setup}
\label{sec:experiments:setup}

We evaluate on ten KTH floor plans and ten publicly available real-SLAM maps, using four starts per map, a 0.1 m grid resolution, and a \(360^\circ\) sensor with ranges \(R\in\{5,10,20\}\) m. All experiments assume exact localization and noise-free sensing. The KTH maps are held-out test maps from the dataset used to train the MapEx ensemble \cite{ho2025mapex}. The real-SLAM set provides a broader geometric test, including two maps that depart substantially from MAPLE's rectilinear prior.
Across all evaluated approaches, frontier candidates are generated from the current partial map using the same procedure, and travel cost is computed as the shortest path through observed free space using A*. MAPLE uses \(N=16\) hypotheses with \(\alpha_f=0.85\), \(\sigma_J=0.35\), \(\rho_s=0.5\), and \(d_s =1.0\) m.

We compare MAPLE against three baselines:  
\begin{itemize}
    \item \emph{\textbf{Nearest:}} classical frontier exploration that selects the frontier candidate with minimum travel cost \cite{yamauchi1997frontier}. 
    \item \emph{\textbf{MapEx:}} a learned LaMa ensemble that predicts occupancy in unknown cells and scores frontier candidates using the ensemble mean and variance \cite{ho2025mapex}. 
    \item \emph{\textbf{IG-Hector:}} a learned map predictor that scores each frontier candidate by the connected unknown area its prediction would reveal \cite{shrestha2019learned}. 
\end{itemize}

We use MapEx’s released LaMa ensemble \cite{suvorov2022resolution, ho2025mapex} and port its published frontier-gain computation into our common framework. For a controlled comparison, MapEx and IG-Hector retain their respective gain estimators while using the common shortest-path travel cost defined in Sec.~\ref{sec:setup}.

We report:
\begin{itemize}
    \item \emph{Exploration performance:} coverage as a function of travel, summarized by the travel required to reach 90\% coverage.
    \item \emph{Prediction quality:} ROC-AUC measures occupied-versus-free discrimination across thresholds, IoU measures overlap between predicted and ground-truth occupied cells, Brier score measures probabilistic prediction error, and NLL penalizes confident prediction errors. Evaluation uses shared partial maps generated by same-range Nearest runs at 20\%, 40\%, and 60\% coverage. Predictions are thresholded at 0.5 for IoU, and MAPLE probabilities are the empirical occupancy frequencies across its map hypotheses. For NLL, probabilities from all methods are clipped to \([10^{-9},1-10^{-9}]\).
    \item \emph{Frontier-decision quality:} one-step decision regret (\(\downarrow\)), measuring the percentage loss in ground-truth gain-to-travel score for the selected frontier candidate relative to the best available candidate.
\end{itemize}

\begin{figure}[t]
  \centering
  \includegraphics[width=\columnwidth]{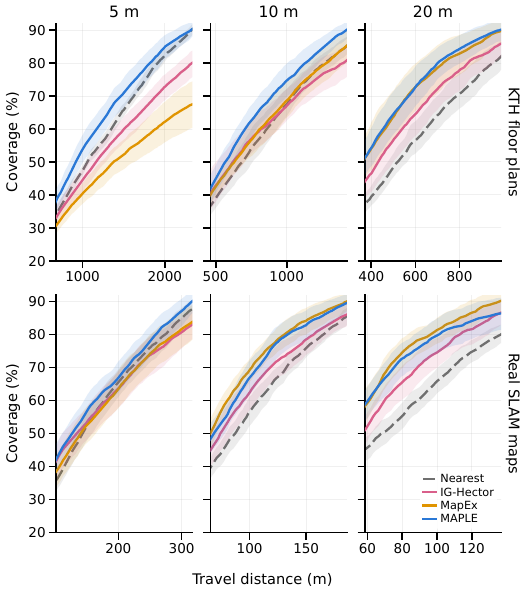}
  \caption{Coverage versus travel distance for KTH floor plans and real-SLAM maps at 5, 10, and 20 m sensor range. Curves show mean coverage over four starts on each of ten maps per family. Shaded regions are 95\% bootstrap intervals clustered by map. Travel is normalized by Nearest's total travel in the corresponding episode and rescaled to the median map in each panel. Curves are shown until the leading method reaches 90\% coverage.}
  \label{fig:coverage}
\end{figure}

\subsection{Exploration Performance}
\label{sec:experiments:exploration}
\begin{figure*}[t]
  \centering
  \includegraphics[width=1\textwidth]{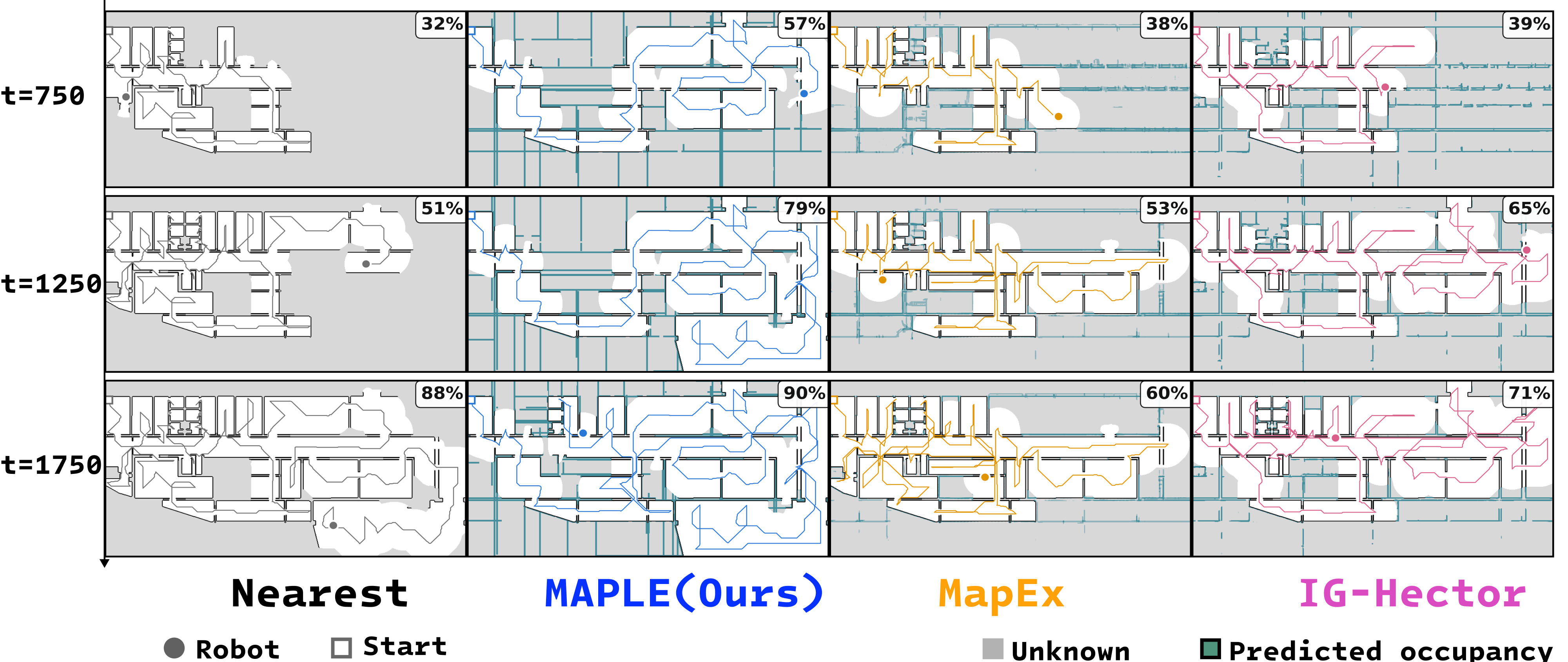}
  \caption{Exploration progression for one KTH floor plan at 5 m sensor range. Columns show Nearest, MAPLE, MapEx, and IG-Hector at equal planning steps. Percentages indicate observed coverage, colored paths show the robot trajectory, and shaded overlays show the current occupancy prediction for the predictive methods.}
  \label{fig:exp_prog}
\end{figure*}

Fig.~\ref{fig:coverage} shows coverage as a function of travel across both map families and all three sensor ranges. MAPLE reaches 90\% coverage with less travel than Nearest and IG-Hector in all six settings. Relative to Nearest, the reduction ranges from 4\% to 23\%, while the reduction relative to IG-Hector ranges from 11\% to 25\%.
The comparison with MapEx depends more strongly on the map family and sensor range. On KTH, MAPLE requires 46\% less travel at 5 m and 18\% less at 10 m, while the two differ by only 1\% at 20 m. On the real-SLAM maps, MAPLE requires 13\% less travel at 5 m but 2\% and 5\% more at 10 m and 20 m, respectively. On the two real-SLAM maps that depart most from the rectilinear prior, MAPLE reduces travel relative to Nearest by 12–32\% across sensor range on Intel Lab. On FR101 it requires 8\% and 5\% more travel at 5 and 10 m but 26\% less at 20 m. MAPLE therefore consistently outperforms Nearest and IG-Hector, while its performance relative to MapEx varies with the map family and sensor range. Fig.~\ref{fig:exp_prog} shows a representative KTH episode at 5 m.

\subsection{Prediction and Frontier-Decision Quality}
\begin{table}[t]
  \centering
  \caption{Prediction quality on unknown cells of shared partial maps, averaged over 20\%, 40\%, and 60\% coverage and maps. Bold marks the best result within each map family and sensor range. Brier and NLL should be compared within map families.}
  \label{tab:prediction}
  \small
  \setlength{\tabcolsep}{3pt}
  \begin{tabular}{lllcccc}
    \toprule
    & Range & Method & ROC-AUC $\uparrow$ & IoU $\uparrow$ & Brier $\downarrow$ & NLL $\downarrow$ \\
    \midrule
    \multirow{9}{*}{\rotatebox{90}{\shortstack{KTH floor plans}}}
    & \multirow{3}{*}{5\,m} & MAPLE & 0.638 & 0.153 & 0.078 & 1.05 \\
    &  & MapEx & \textbf{0.696} & 0.189 & \textbf{0.069} & \textbf{0.85} \\
    &  & IG-Hector & 0.677 & \textbf{0.209} & 0.081 & 1.23 \\
    \cmidrule(l){2-7}
    & \multirow{3}{*}{10\,m} & MAPLE & 0.636 & 0.152 & 0.076 & 1.01 \\
    &  & MapEx & \textbf{0.679} & 0.160 & \textbf{0.070} & \textbf{0.86} \\
    &  & IG-Hector & 0.668 & \textbf{0.197} & 0.080 & 1.21 \\
    \cmidrule(l){2-7}
    & \multirow{3}{*}{20\,m} & MAPLE & 0.630 & 0.154 & 0.074 & 0.99 \\
    &  & MapEx & \textbf{0.678} & 0.167 & \textbf{0.069} & \textbf{0.84} \\
    &  & IG-Hector & 0.662 & \textbf{0.192} & 0.079 & 1.18 \\
    \midrule
    \multirow{9}{*}{\rotatebox{90}{\shortstack{Real SLAM maps}}}
    & \multirow{3}{*}{5\,m} & MAPLE & 0.636 & \textbf{0.190} & \textbf{0.156} & 2.16 \\
    &  & MapEx & \textbf{0.671} & 0.127 & 0.165 & \textbf{2.04} \\
    &  & IG-Hector & 0.658 & 0.145 & 0.192 & 2.96 \\
    \cmidrule(l){2-7}
    & \multirow{3}{*}{10\,m} & MAPLE & 0.630 & \textbf{0.181} & \textbf{0.158} & 2.14 \\
    &  & MapEx & \textbf{0.657} & 0.117 & 0.168 & \textbf{2.13} \\
    &  & IG-Hector & 0.650 & 0.136 & 0.195 & 3.00 \\
    \cmidrule(l){2-7}
    & \multirow{3}{*}{20\,m} & MAPLE & 0.632 & \textbf{0.185} & \textbf{0.160} & 2.22 \\
    &  & MapEx & \textbf{0.669} & 0.131 & 0.169 & \textbf{2.14} \\
    &  & IG-Hector & 0.656 & 0.142 & 0.199 & 3.10 \\
    \bottomrule
  \end{tabular}
\end{table}

The exploration results do not reveal whether performance comes from more accurate occupancy prediction or from how predictions are used to select frontier candidates. Table~\ref{tab:prediction} therefore compares prediction quality on the same partial maps. On KTH, MapEx is the strongest cell-level predictor, leading ROC-AUC, Brier, and NLL at every range, while IG-Hector gives the highest IoU. On the real-SLAM maps, the ordering is mixed. MapEx retains the best ROC-AUC and NLL, while MAPLE gives the best IoU and Brier. Prediction quality alone therefore does not reproduce the exploration ordering.
\begin{figure}[t]
  \centering
  \includegraphics[width=\columnwidth]{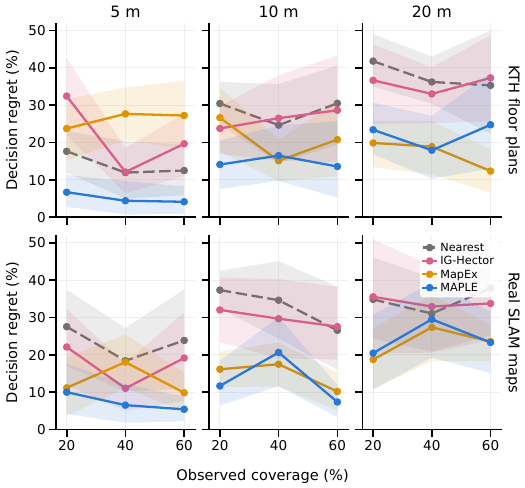}
  \caption{One-step decision regret on the shared partial maps. Rows show map family and columns sensor range. Regret is the percentage loss in ground-truth gain-to-travel score relative to the best available frontier candidate, so lower is better. Shaded regions are 95\% bootstrap intervals clustered by map.}
  \label{fig:regret}
\end{figure}

Fig.~\ref{fig:regret} evaluates the resulting frontier decisions on those same states using one-step decision regret. Across the 18 conditions defined by map family, sensor range, and observed coverage, MAPLE has the lowest regret in 12. Its average regret is 14.5\%, compared with 19.2\% for MapEx, 27.4\% for IG-Hector, and 28.5\% for Nearest. The advantage is strongest at short range, particularly on KTH at 5 m, where MAPLE loses only 4--7\% of the best available gain-to-travel score across the three coverage levels, compared with 24--28\% for MapEx.
MapEx leads in the remaining 6 conditions, mostly at 20 m, and many individual confidence intervals overlap. Taken together with Table~\ref{tab:prediction}, these results show that the ordering in cell-level prediction does not directly determine the ordering in frontier decisions. This mismatch is strongest at shorter sensor ranges, where more of each decision depends on unseen space.

\subsection{Ablation Studies}
\label{sec:experiments:ablations}

\begin{figure}[t]
  \centering
  \includegraphics[width=0.95\columnwidth]{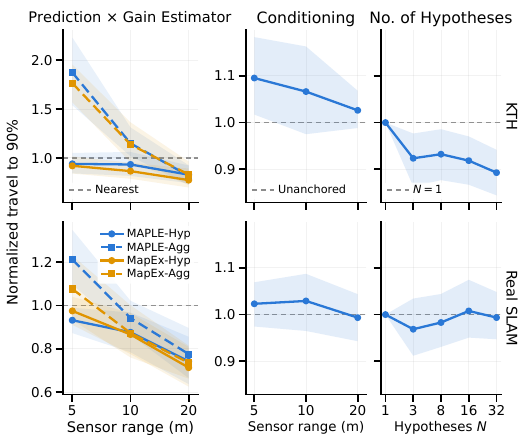}
    \caption{Ablation results on KTH and real-SLAM maps. Left compares the MAPLE and MapEx map-prediction methods under hypothesis-wise and aggregated gain estimation with \(N=3\). Middle compares MAPLE with its unanchored variant that removes conditioning on observed geometry. Right varies the number of MAPLE hypotheses at 20 m range relative to \(N=1\). Travel to 90\% coverage is normalized to Nearest (left), unanchored MAPLE (middle), and \(N=1\) (right), with lower values indicating better exploration efficiency. Shaded regions show 95\% bootstrap intervals clustered by map.}
  \label{fig:ablation}
\end{figure}

We next use controlled ablations to separate the effects of the map-prediction method, gain estimation, geometric conditioning, and number of hypotheses on exploration performance.

Fig.~\ref{fig:ablation} first separates the effect of the map-prediction method and gain estimator, with MAPLE and MapEx each using \(N = 3\) predictions. For MapEx-Hyp, each LaMa prediction is thresholded at 0.5 over \(U_t\) to form a binary map hypothesis evaluated with Eq.~(\ref{eq:igps}). MapEx-Agg retains its published mean-and-variance estimator. On KTH at 5 m, hypothesis-wise gain estimation reduces travel by 49.9\% for MAPLE and 47.9\% for MapEx relative to aggregation. The advantage decreases with sensor range and is weaker on real-SLAM. In most conditions, changing the map-prediction method while holding the gain estimator fixed has a smaller effect. This indicates that at short range, especially on KTH, how the predictions are evaluated matters more than the map-prediction method itself.

We next isolate the effect of MAPLE's conditioning on observed geometry. Removing this conditioning does not consistently increase exploration travel. On KTH, the conditioned prior requires 9.5\% more travel at 5 m and 2.6\% more at 20 m, while real-SLAM differences remain within 3\% with overlapping intervals. Conditioning nevertheless changes prediction quality substantially, raising ROC-AUC from about 0.47 to 0.63–0.64 and IoU from 0 to 0.15–0.19 on the same states, although Brier and NLL do not improve consistently. Thus, conditioning substantially improves MAPLE's map-prediction quality, but the corresponding exploration travel does not improve consistently. This further shows that improved cell-level prediction quality does not necessarily translate into more efficient exploration.

Finally, we vary the number of hypotheses. At 20 m on KTH, \(N=3\) already reduces travel by 7.6\% relative to \(N=1\), compared with 10.7\% at \(N=32\). Real-SLAM shows no clear improvement with more hypotheses. MAPLE therefore obtains most of its benefit with only a few hypotheses.

\subsection{Computational Cost}
\label{sec:experiment:compute}
With \(N=16\), a complete MAPLE replanning step takes a median of 4.00 s on KTH and 1.72 s on real-SLAM using one Intel Core Ultra 9 185H CPU core. Under the same CPU constraint, MapEx requires 76.5 s and 38.7 s, while IG-Hector requires 25.6 s and 13.0 s. MAPLE is therefore 19–23× and 6.4–7.5× faster, respectively. On an NVIDIA RTX 4050, MapEx drops to 2.65 s and 1.17 s and IG-Hector to 0.93 s and 0.39 s. Thus, the learned baselines are faster with GPU acceleration, while MAPLE remains practical without learned inference or dedicated GPU hardware.


\section{Conclusion}
\label{sec:conclusion}

We introduced MAPLE, a training-free generative prior for predictive indoor exploration that samples multiple plausible map hypotheses and evaluates frontier visibility separately within each hypothesis before averaging the resulting gains. Across KTH and real-SLAM environments, MAPLE consistently outperforms non-predictive exploration and remains competitive with learned predictive methods while operating on a single CPU core.
Beyond end-to-end performance, the results show that cell-level prediction accuracy does not directly determine exploration quality. How map predictions are represented and evaluated also matters. The ablations reinforce this finding and show that a few map hypotheses capture most of MAPLE's benefit.

Although the Mondrian process extends to higher dimensions, the current MAPLE formulation and evaluation are two-dimensional, motivating future extension of its geometric conditioning to 3D occupancy maps. Future work will also test MAPLE on physical robots under sensing noise and SLAM drift, and in environments that depart more strongly from its rectilinear prior. We also plan to extend MAPLE's multi-hypothesis representation to indoor target search, where each map hypothesis can support reasoning about both unseen space and possible target locations.

\bibliographystyle{IEEEtran}
\bibliography{references}

\end{document}